\pdfoutput=1

\documentclass[journal]{IEEEtran}

\usepackage[numbers]{natbib}
\usepackage{graphicx} 
\graphicspath{{./Figures/}}               
\usepackage{color}
\usepackage{subfig}
\usepackage{amssymb, amsmath}                            
\usepackage{amsfonts}
\usepackage{mathtools}

\usepackage{tabularx}
\usepackage{booktabs}
\usepackage{colortbl} 
\usepackage{xcolor} 
\usepackage{xfrac}
\usepackage{multirow}

\usepackage[linesnumbered,ruled]{algorithm2e}
\usepackage{hyperref}

\begin{document}

\title{Uncertainty-aware Self-supervised Learning for Cross-domain Technical Skill Assessment in Robot-assisted Surgery}

\author{Ziheng Wang\thanks{Ziheng Wang was with
              Department of Mechanical Engineering, The University of Texas at Dallas, Texas, USA.
              email: zihengwang@intusurg.com}, Andrea Mariani\thanks{Andrea Mariani is with
              The BioRobotics Institute and the Department of Excellence in Robotics \& AI, Scuola Superiore Sant'Anna, Pisa, Italy.}, Arianna Menciassi\thanks{Arianna Menciassi is with
              The BioRobotics Institute and the Department of Excellence in Robotics \& AI, Scuola Superiore Sant'Anna, Pisa, Italy.}, Elena De Momi\thanks{Elena De Momi is with the
              Electronic Information and Bioengineering Department (DEIB), Politecnico di Milano, Italy.}, Ann Majewicz Fey\thanks{Ann Majewicz Fey is with
              Department of Mechanical Engineering, The University of Texas at Austin \&
              Department of Surgery, UT Southwestern Medical Center, Texas, USA.}}

\maketitle
\begin{abstract} 

Objective technical skill assessment is crucial for effective training of new surgeons in robot-assisted surgery. With advancements in surgical training programs in both physical and virtual environments, it is imperative to develop generalizable methods for automatically assessing skills. In this paper, we propose a novel approach for skill assessment by transferring domain knowledge from labeled kinematic data to unlabeled data. Our approach leverages labeled data from common surgical training tasks such as \textit{Suturing}, \textit{Needle Passing}, and \textit{Knot Tying} to jointly train a model with both labeled and unlabeled data. Pseudo labels are generated for the unlabeled data through an iterative manner that incorporates uncertainty estimation to ensure accurate labeling. We evaluate our method on a virtual reality simulated training task (Ring Transfer) using data from the da Vinci Research Kit (dVRK).
The results show that trainees with robotic assistance have significantly higher expert probability compared to these without any assistance, $p < 0.05$, which aligns with previous studies showing the benefits of robotic assistance in improving training proficiency.
Our method offers a significant advantage over other existing works as it does not require manual labeling or prior knowledge of the surgical training task for robot-assisted surgery.
\end{abstract}

\begin{IEEEkeywords}
Surgical skill assessment, surgical training, Bayesian deep learning, virtual reality, robot-assisted surgery
\end{IEEEkeywords}

\section{Introduction}
Robot-assisted minimally-invasive surgery (RAMIS) with the advance of robotic platforms, such as the \textit{da Vinci} surgical system (Intuitive Surgical Inc., Sunnyvale, CA), has revolutionized surgical interventions towards a safe, precise, and less invasive approach for patient care~\cite{lanfranco2004robotic,beasley2012medical}.
Despite benefits, robot-assisted surgery is technically complex and different for clinical operators compared to traditional open interventions~\cite{talamini2003prospective,weber2018effects}. It requires trainees to develop fundamental technical skills, map task perception to appropriate actions, and take valid operational strategies to properly control the surgical tools~\cite{okamura2009haptic}. 
Such proficiency could only be obtained by sufficient training and structured practice with appropriate feedback~\cite{vedula2017objective}. 
Therefore, understanding technical skills in various scenarios and providing accurate assessment for medical trainees has paramount relevance in the field~\cite{maier2018surgical}.


Typically, the acquisition of technical skills for surgery is supported by verbal feedback from senior surgeons~\cite{darzi1999assessing,kotsis2013application}, and assessed using methods that are largely subjective and resource-expensive, such as structured checklists and rating scales~\cite{aghazadeh2015GEARS,hatala2015constructing}. 
Recent advent of surgical robotic platforms and Virtual Reality (VR) or computer-based simulators allows to collect a broad set of recordings that are relevant to various surgical operations~\cite{d2021accelerating,kumar2015current,lerner2010does,abboudi2013current,albani2007virtual,al2008role}. Together with the advancements in surgical data science, there is an increasing interest to assess technical skills using tool kinematics and endoscopic videos collected during basic training tasks~\cite{maier2017surgical,nagy2017surgical}.
In addition, recent studies in training augmentation have shifted the focus towards the development of adaptive strategies for personalized, self-directed surgical training~\cite{macdonald2003self, siu2016adaptive,enayati2018robotic,zhang2019self}. This approach is driven by the goal of maximizing learning by adapting some training parameters to the trainee. More specifically, this adaptation relies on a quantification of the trainee's skill where a training curricula or robotic assistance is adapted accordingly as a function of the trainee's learning progress~\cite{enayati2018robotic, mariani2018design, mariani2020skill, zhang2019self}. Thus, an accurate and continuous skill assessment is highly relevant to enable an appropriate adaptation criterion and to provide adequate feedback and timely guidance. 
Taken together, these factors lead to a distinct necessity of an efficient solution for assessing skills in personalized surgical training to better train future robotic surgeons.

Pioneering surgical skill models ranging from shallow classifiers to advanced deep networks have been broadly exploited. Classical machine learning algorithms such as support vector machine (SVM), linear discriminant analysis (LDA), and hidden Markov models (HMM) are widely used to assess skills based on low-level features extracted from either videos or kinematic data~\cite{kassahun2016surgical, fard2018automated, lin2006towards, chmarra2010objective,tao2012sparse}. In contrast, recent development of deep learning leads to an increasing popularity due to its superior end-to-end learning and automatic feature extraction capabilities. With a sufficient amount of labeled data, convolutional neural network (CNN) and its modifications could provide a high classification accuracy in several basic surgical training exercises~\cite{wang2018deep, jin2018tool, funke2019video, fawaz2019accurate, kim2019objective}. 
As sequence learners, recurrent neural networks (RNN) such as long-term-short-memory (LSTM) could explore complex temporal dynamics of the sequential data and have shown a competitive accuracy for measuring skills from motion kinematics~\cite{wang2018satr,nguyen2019surgical}.

\begin{figure*}[t]
      \centering
      \includegraphics[width=1.00\linewidth]{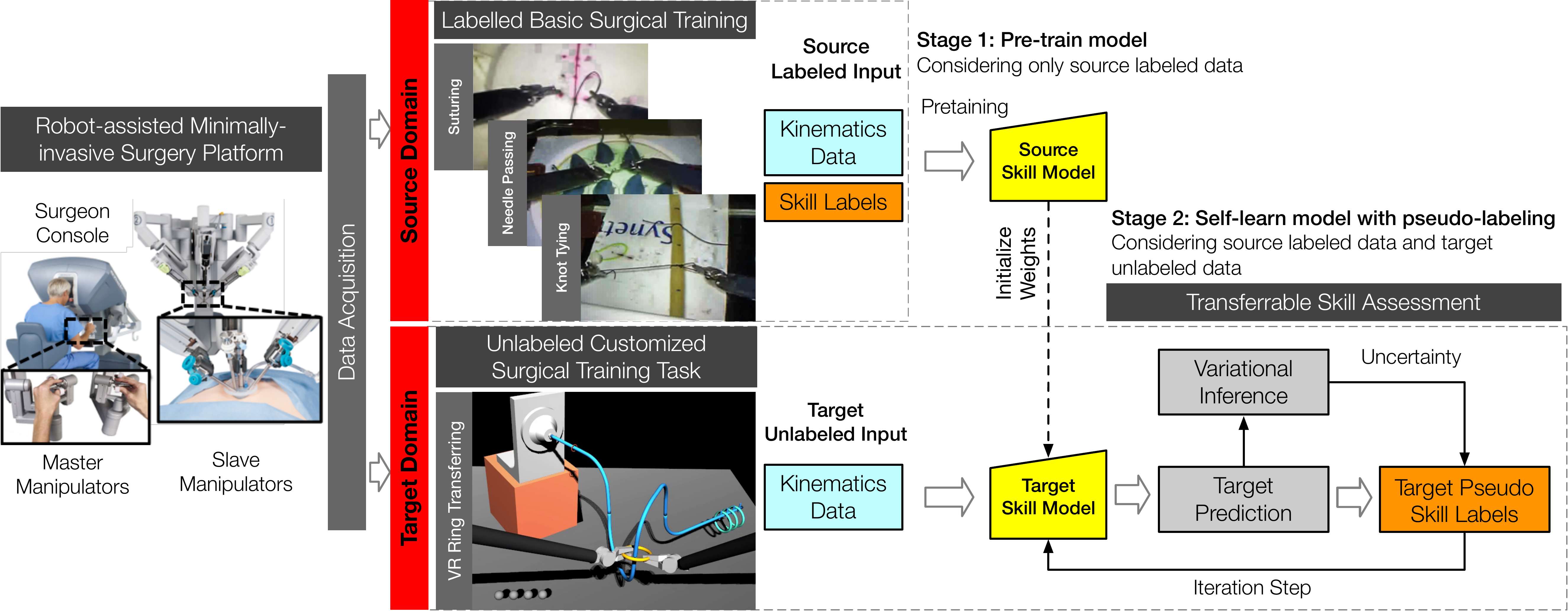}
      \caption{Cross-domain technical skill assessment for robot-assisted surgery. Our proposed method starts with pre-training a model using only the labeled kinematics data from a source domain, i.e., established common surgical training exercises (referred to as source domain data). This pre-trained model acts as a "teacher" to understand the unlabeled surgical training tasks of interest in the target domain. Our approach leverages both the labeled source domain data and the abundant unlabeled data from the target domain to improve representation learning. With the integration of an uncertainty estimation, the resulting skill model is able to capture informative, domain-invariant knowledge from kinematics data. No manual labeling of surgical training tasks in the target domain is necessary.}
      \label{fig_flow}
\end{figure*}

In general, there are two main drawbacks that existing approaches typically feature when building a predictive model for skill assessment, particularly given broader, personalized surgical training exercises. 
First, all these methods only utilize the labeled data and are developed to assess skills of specific tasks in surgical training, such as \textit{Suturing}, \textit{Needle Passing}, and \textit{Knot Tying}~\cite{gao2014jhu}.  Despite high performance of skill assessment has been reported in these studies, their effectiveness in generalizing to new unlabeled data is less explored. 
In particular, the abundant unlabeled data from the complex settings of surgical training tasks can be very different from the labeled ones. It remains less known whether the model can generate meaningful skill labels given the new settings of surgical training tasks (i.e. assessing skills in a training task that is very different from the ones used to train the skill model).
Secondly, most existing methods assume the availability of skill labels as the ground-truth required for modeling. However, manual labeling is time-consuming and labor-intensive as it involves significant efforts of experienced professionals to assess skills based on their case review~\cite{gao2014jhu}. A new approach that can jointly explore the intrinsic characteristics of unlabeled surgical data while leveraging available information of labeled data is thus desirable.

This work focuses on the cross-domain assessment of technical skills in surgical training tasks based on kinematic data. We present a two-stage framework that incorporates self-supervised learning to capture inherent knowledge from both labeled and unlabeled data. An uncertainty estimation is used to generate appropriate pseudo labels for the target domain. Our approach leverages both labeled and unlabeled data, allowing the model to capture more inherent features of kinematics that are less specific to a particular domain. 
Additionally, our approach does not require prior skill labeling in the target domain, enabling the assessment of skills without the constraint of labeling which is typically challenging to acquire. 

The key contributions of this paper are:
\begin{itemize}
\item A novel approach to assess surgical skills in unlabeled kinematic data for new robotic surgical training exercises.
\item A Bayesian self-supervised model with uncertainty regularization that learns domain-invariant features from both labeled and abundant unlabeled data.
\item Validation of our approach on an unlabeled dataset from standalone virtual reality simulation-based surgical training. By using pseudo labeling, we are able to extract trainees' learning curves and demonstrate alignment with performance measures in VR training.
\item A data-driven, generic approach that eliminates the need for manual labeling and prior task-specific knowledge for heterogeneous surgical training exercises.
 
\end{itemize}


\section{Methodology} 
We aim to assess technical skills in a unlabeled surgical training task. We explicitly focus on a personalized, non-standard training task, VR simulated \textit{Ring Transfer}, which is different from a typical setting of basic surgical training. We formalize the problem as a cross-domain adaptation question for skill assessment. Differently from conventional fully supervised (only trained on labeled data) and fully unsupervised learning (trained on unlabeled data) paradigms, our approach seeks to learn a model by generalizing a source for which we have labeled kinematic data with ground-truth skill labels to a target domain where no true target skill label is available. 

\subsection{Preliminary}
Formally, we denote the source domain as a group of labeled samples, $\mathcal{D}_L:= \{(x_1, y_1), \cdots, (x_i, y_i) \} = \{(X, Y)\}$, that were collected from basic skill training. Each $i$-th sample pair is composed by the input kinematics data $x_i \in \mathbb{R}^d$, associated with a corresponding skill class label $y_i \in \{1, \cdots, K\}$. Here, $d$ is the dimension of input data and $K$ is the number of classes. 
Similarly, we denote the target domain as a set of unlabeled samples $\mathcal{D}_U:= \{x_1, \cdots, x_j\} = \{X^\prime\}$, $x_j\in \mathbb{R}^d$. Target domain was collected from a previously unseen exercise involving robotic surgical skills, where no ground truth labels of target domain are available.

In general, domain adaption refers to a set of approaches predicting the labels of samples from target domain $\mathcal{D}_U$, using both labeled samples from source domain and unlabeled samples from the target domain itself. As follows, the goal of transferable skill assessment is to train a model $f_{\theta}(\cdot)$ overs $\mathcal{D}_L$ and $\mathcal{D}_U$ that enables learning of both transferable features and domain-invariant classifier. Here,  the output of $f_{\theta}(\cdot)$ is the predictive class probability of skills and $\theta$ denotes the learned network parameters.

\subsection{Self-supervised Learning Protocol}
Self-supervised learning, also known as self-learning, refers to a protocol that is specifically designed to learn from both labeled and unlabeled data~\cite{zhu2005semi}. Typically, self-learning approaches first construct an initial classifier based on the labeled data. Then, the classifier is applied to some of unlabeled data and make predictions using the model. These predictions, namely $pseudo$ labels, are included to re-train the existing classifier in an iterative fashion until a certain stopping criterion is satisfied. 

We set up a similar procedure and devoted to domain adaptation for transferable skill assessment: train a classifier on the labeled source domain, produce approximate \textit{pseudo} labels of surgical exercises in target domain. Next, these \textit{pseudo}-labeled target samples are treated as labeled ones and used for re-training a new classifier in target domain. Then, the classifier explores the remaining set of target domain until model learning converges or the maximum iteration number is reached. 
In essence, self-learning considers the information from both domains and enables to propagate knowledge from the labeled source to the unlabeled target. Therefore, it could not only exploit the hidden structures in motion kinematic data that regularize learning, but help to find a shared feature space that matches data distributions of both the domains. 

We train the whole network (with parameters $\theta$) by minimizing the loss between the model output $f_{\theta}(x)$ and the labels (true labels from source and $pseudo$ labels from target), while penalizing large activation.
Here, $y$ denotes the true labels in source domain $x \in \mathcal{D}_{L}$, and $y^\prime$ denotes the \textit{pseudo} labels of unlabeled samples in target domain $x \in \mathcal{D}_{U}$, 
Thus, the overall loss $\mathcal{L(\theta)}$ is a weighted mixture of the cross-entropy loss of labeled source domain, $\mathcal{L}_{labeled}$, the cross-entropy loss of pseudo-labeled ones from target domain, $\mathcal{L}_{pseudo}$, and a $L2$ weight regularizer term:

\begin{equation}
\begin{aligned}
\mathcal{L} & = \mathcal{L}_{labeled}+ \alpha \mathcal{L}_{pseudo} + \lambda \lVert\theta\rVert^2\\
&     = \sum\nolimits_{x \in\mathcal{D}_{L}} \mathcal{L}(y, f_\theta(x)) + \alpha \sum\nolimits_{x \in \mathcal{D}_{U}} \mathcal{L}(y^\prime, f_\theta(x)) + \lambda \lVert\theta\rVert^2
\end{aligned}
\end{equation}
where $\alpha$ and $\lambda$ are two non-negative hyper-parameters to balance contributions of three loss components. In particular, the parameter $\alpha$ controls the importance of $pseudo$-labeled target samples relative to the true labeled source samples. A larger $\alpha$ could encourage selecting more $pseudo$-labels for training, whereas a smaller $\alpha$ reduces the impact of target samples for training.

 \begin{algorithm}
    \caption{\bf{Technical skill assessment with self-learning and uncertainty estimation.}}
    \label{algorithm1}
    \SetKwInOut{Input}{Input}
    \SetKwInOut{Output}{Output}
    \SetKwInOut{Initialization}{Initialization}
    \SetKwInOut{return}{return}
    \Input{Labeled source domain $\mathcal{D}_L$, unlabeled target domain $\mathcal{D}_U$, maximum iteration $N$, inference round $T$, learning rate $\alpha$, pre-trained parameters $\theta^\prime$}
    \Output{Optimized network parameters $\theta$}
    \Initialization{Let training set $\mathcal{D}_{train}$ = $\mathcal{D}_L$, initialize parameters with pre-trained network $\theta^\prime$, iteration index $n=1$}
    \Repeat{\textit{converge} or $n==N$} 
      {       
      \For{all $x \in \mathcal{D}_U$} {Variational MC-dropout inference with $T$ stochastic forward passes\\
      Determine uncertainty of the input sample $x$
      }
      Infer the softmax class probability of $pseudo$ labels for unlabeled target samples $y^\prime$\\
      Select the first $k$ target samples with lowest uncertainty of $pseudo$ labels, $\mathcal{D}_U^{\prime}$\\
      Update training set by $\mathcal{D}_{train} \leftarrow \mathcal{D}_{train} \cup \mathcal{D}_U^{\prime}$\\
      Update remaining unlabeled dataset in the target domain by $\mathcal{D}_U \leftarrow \mathcal{D}_U - \mathcal{D}_U^\prime$\\
      \If{$n \% 2==0$} 
      {Update a learning rate by $\alpha^{n}  \leftarrow \alpha^{n-1}*0.5$}
      Feed forward the current training data $\mathcal{D}_{train}$, and backward propagation minimizing cost $\mathcal{L}{(\theta)}$\\
      Update the network parameters $\theta$ \\
     $n \leftarrow n+ 1$\\
      }
      {return $\theta$}
\end{algorithm}

\subsection{Uncertainty Estimation}
Jointly learning classifiers and optimizing \textit{pseudo} labels on unlabeled target domain is naturally difficult since it is not possible to completely guarantee the correctness of generated \textit{pseudo} labels given an existing model. The uncertainty could come from the variability in the relationship between observed tool kinematics data and the corresponding skill class outputs. 
Conventionally, $pseudo$ labels were selected for self-learning by taking the class which has the maximum softmax class probability or the one with corresponding probability that is larger than a predefined threshold (chosen experimentally). The softmax function approximates relative class probabilities, but it did not capture any uncertainty or confidence information regarding the prediction outputs~\cite{gal2016uncertainty}.
To cope with this issue, our strategy is first to estimate the model uncertainty of skill predictions and then seek for \textit{pseudo} labels from the most confident ones in unlabeled target domain. These high-confidence \textit{pseudo} labels presumably approximate the underlying true target labels and could thus better help self-learning adaptation. 

Bayesian probability theory has offered us a statistically principled way to infer the uncertainty within deep learning tools~\cite{gal2015bayesian}. Research has shown that any deep learning model can be arbitrarily interpreted into Bayesian ones, which are often referred as Bayesian neural networks.
In Bayesian settings, the model uncertainty, also referred to as epistemic uncertainty, could be readily derived from the variances of prediction with respect to the distribution of network parameters $\theta$, i.e., posterior $p(\theta |X, Y)$. The distribution characterizes the uncertainty or confidence of model that conforms to the observed data $(X, Y)\in\mathcal{D}_{train}$.
Although the posterior $p(\theta |X, Y)$ is analytically intractable, a popular approach, variational inference, can approximate it with a simpler approximating distribution that is much easier to sample with, while minimizing the Kullback-Leibler (KL) divergence between the approximating distribution and the true posterior~\cite{gal2016dropout}. Kendal \emph{et al.}~\cite{kendall2017uncertainties} show that the variational inference using stochastic Monte-Carlo dropout (MC dropout) could allow to approximate Bayesian posterior and minimize the cross-entropy loss of a network with dropout is equivalent to minimizing the KL divergence. Differently from the standard dropout that is used for regularization in training, MC approach stochastically drops out parameters of existing model at test time, which is equivalent to impose a Bernoulli distribution on parameters, to obtain the posterior approximate.

To allow uncertainty estimate for self-learning, we extend and cast our skill model into the Bayesian setting and implement MC dropout to obtain the uncertainty estimates over the current network. In practice, estimating prediction uncertainty was accomplished as follows: (1) Given an existing classifier with network parameters $\theta$, we run the network with a chosen number of stochastic forward passes $T$, where each unit has a probability $p$ of being set to 0.
(2) We collect the softmax probability outputs for each skill class in the $t$-th forward pass ($t=1,\cdots,T$). (3) From the outputs over $T$ forward passes, the final prediction of class probability and uncertainty are taken to be the average and variance of predictions of each class, as defined in Eq.~\ref{eq_average} and Eq.~\ref{eq_uncertainty}, respectively. Further, we derive the entropy of per-class uncertainty as the overall uncertainty measure given the input data. 

\begin{equation}
\label{eq_average}
y^\prime \simeq \frac{1}{T}\sum\limits_{t=1}^{T}{y_t^\prime} =  \frac{1}{T}\sum\limits_{t=1}^{T}{f_{\hat\theta}(x)}  \ \ \forall x \in X^\prime 
\end{equation}

\begin{equation}
\label{eq_uncertainty}
Var(y^\prime) =\frac{1}{T}\sum\limits_{t=1}^{T} (y_t^\prime - y^\prime)^2 
\end{equation}
where $f_{\hat\theta}(\cdot)$ denote $f_\theta(\cdot)$ when including MC dropout, $\hat\theta$ is the current network parameter $\theta$ masked by the dropout during inference.

Taken together, Algorithm~\ref{algorithm1} depicts the overall procedure in details. Initially, the inputs are the labeled samples from source domain $\mathcal{D}_{L}$ and unlabeled samples of target domain $\mathcal{D}_{U}$. After initialization with pre-trained network parameters $\theta^\prime$, our approach performs the following major steps:
(1) Given current model parameters, MC-dropout variational inference was conducted to query the uncertainty of predictions over the unlabeled samples $x\in\mathcal{D}_{U}$.
(2) Each unlabeled sample are assigned with $pseudo$ labels by taking the class which has the maximum average class probability. 
(3) Subset of target domain $\mathcal{D}_U^{\prime} \subset \mathcal{D}_U$ in which the target samples with the first $k$ lowest amounts of uncertainty are added to the current training set $\mathcal{D}_{train} \leftarrow \mathcal{D}_{train} \cup \mathcal{D}_U^{\prime}$ and simultaneously removed from unlabeled target domain $\mathcal{D}_U \leftarrow \mathcal{D}_U - \mathcal{D}_U^\prime$.  
(3) Re-train a new classifier using the updated training set $D_{train}$, and update network parameters $\theta$.
The classifier explores the remaining unlabeled data $\mathcal{D}_{U}$ with smaller confidence until learning converges or maximum iteration is reached. 

In our experiment, we fixed MC-dropout probability as $p=0.5$ for each dropout layer, the total number of forward passes for inference was set as $T=50$ as we sample $T=50$ prediction outputs for each input, and the number of selected target samples ($pseudo$-labeled) in each training iteration was $k=50$. The hyper-parameters were chosen empirically as they have shown a faster training convergence for modeling.
As our model may no longer achieve the best classification for source domain after adapting to the target domain, we adopt the term "converge" to denote the best model for which it is confident enough to make a prediction for the target domain.

\begin{figure}[t]
      \centering
      \includegraphics[width= 1.00\linewidth,clip ,trim=0pt 0pt 0pt 0pt]{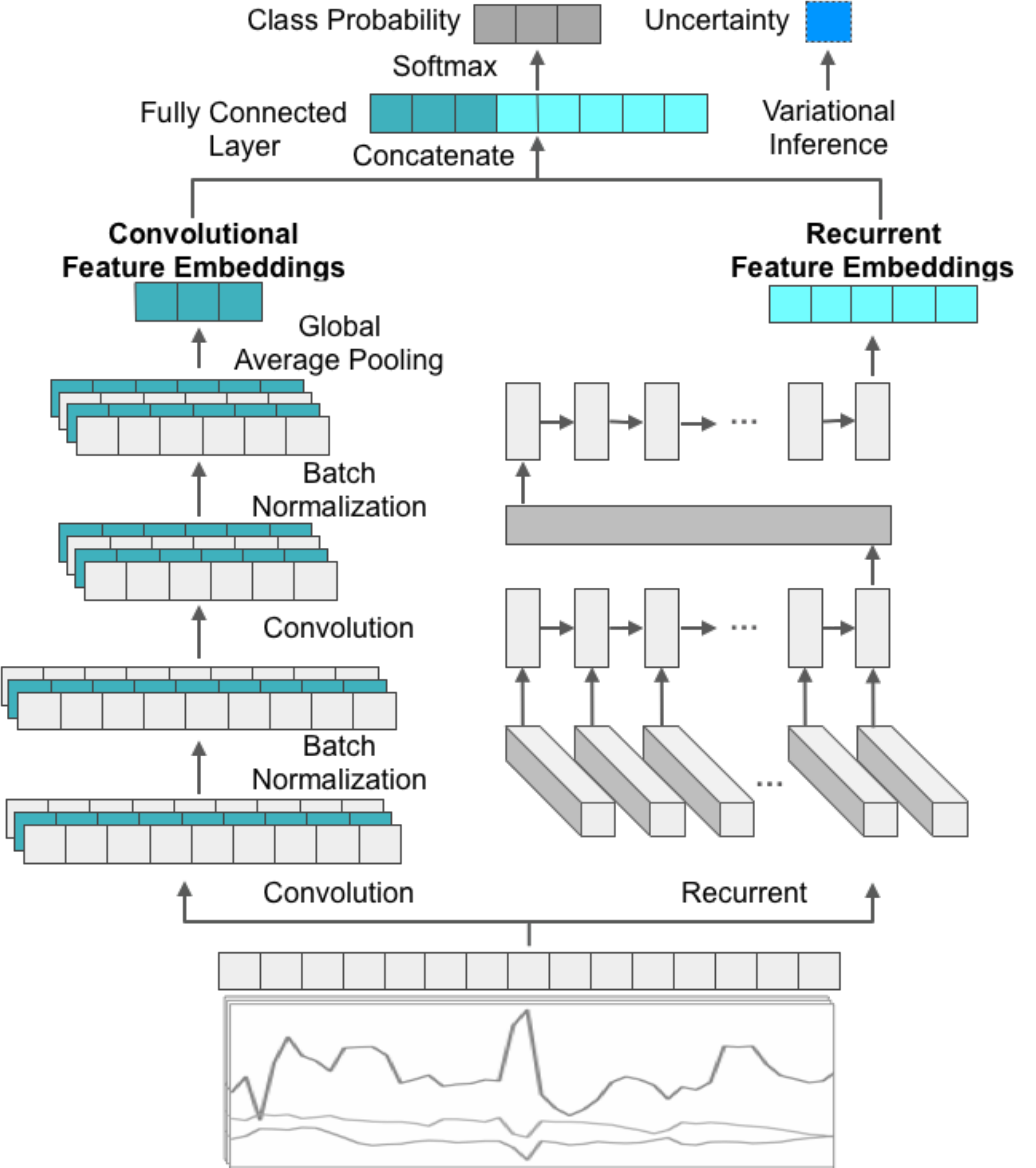}
      \caption{Overview of the proposed architecture for technical skill assessment. The network utilizes multi-dimensional tool kinematics data collected from da Vinci surgical robot API. The output is the softmax class probability of skills along with an uncertainty estimate of the predicted skill.}
      \label{fig_framework}
\end{figure}

\subsection{Architecture}
Fig~\ref{fig_framework} shows the overview of our proposed architecture. The network input is the motion kinematics measured from robot end-effectors and the output is an estimated softmax probability of skill classes that the input data is drawn from, along with an estimate of prediction uncertainty.

The network consists of two main components where the input data flows through in parallel: a convolutional component and a recurrent component. The convolution component comprises two stacked convolutional layers that extract local patterns along the length of time-series data. Each convolutional layer is followed by a ReLU activation and a standard dropout layer with 0.2 dropout rate to reduce the risk of overfitting. A global average pooling layer is then applied to sum out the spatial information. It enables the network to handle input sequences of varying lengths. Also, the global average pooling could largely reduce the number of network parameters and thus help to alleviate overfitting. In parallel, the recurrent component consisted of two bi-directional LSTMs followed by a standard dropout layer. The two sets of features from both components are concatenated and then fed through a dense layer. Finally, the network is forked at the last layer to have two outputs: the skill probability output through a layer with softmax nonlinearity, and an uncertainty estimate obtained from the aforementioned Bayesian variational inference.

\begin{figure}[tb]
      \centering
      \includegraphics[width= 0.9\linewidth,clip ,trim=0pt 0pt 0pt 0pt]{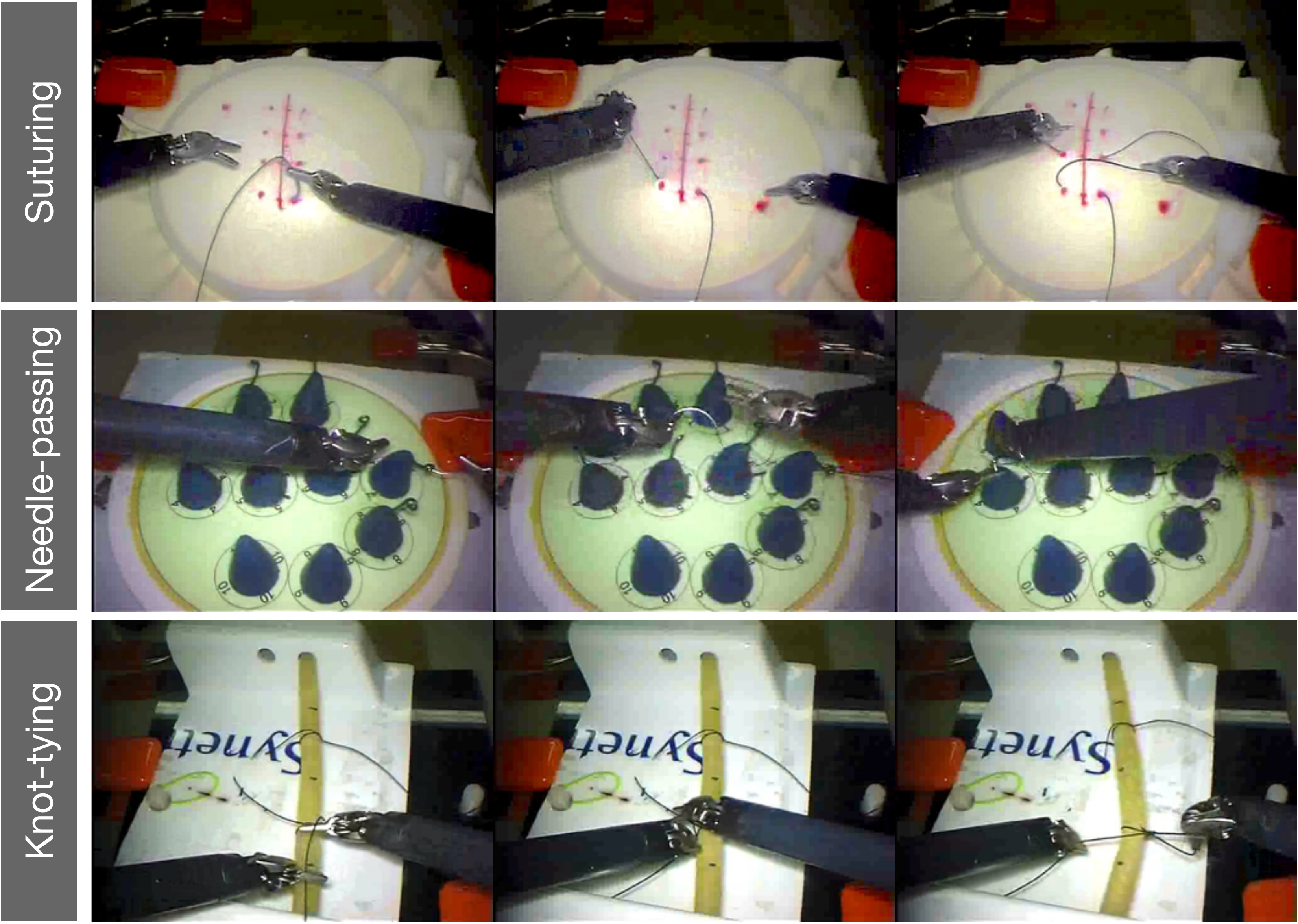}
      \caption{Snapshots of trainee operations in three basic surgical training tasks (source domain dataset), including \textit{Suturing}, \textit{Needle Passing}, and \textit{Knot Tying}~\cite{gao2014jhu}.}
      \label{fig_jigsaws}
\end{figure}

 \begin{figure*}[tb]
      \centering
      \includegraphics[width= 0.95\linewidth,clip ,trim=0pt 0pt 0pt 0pt]{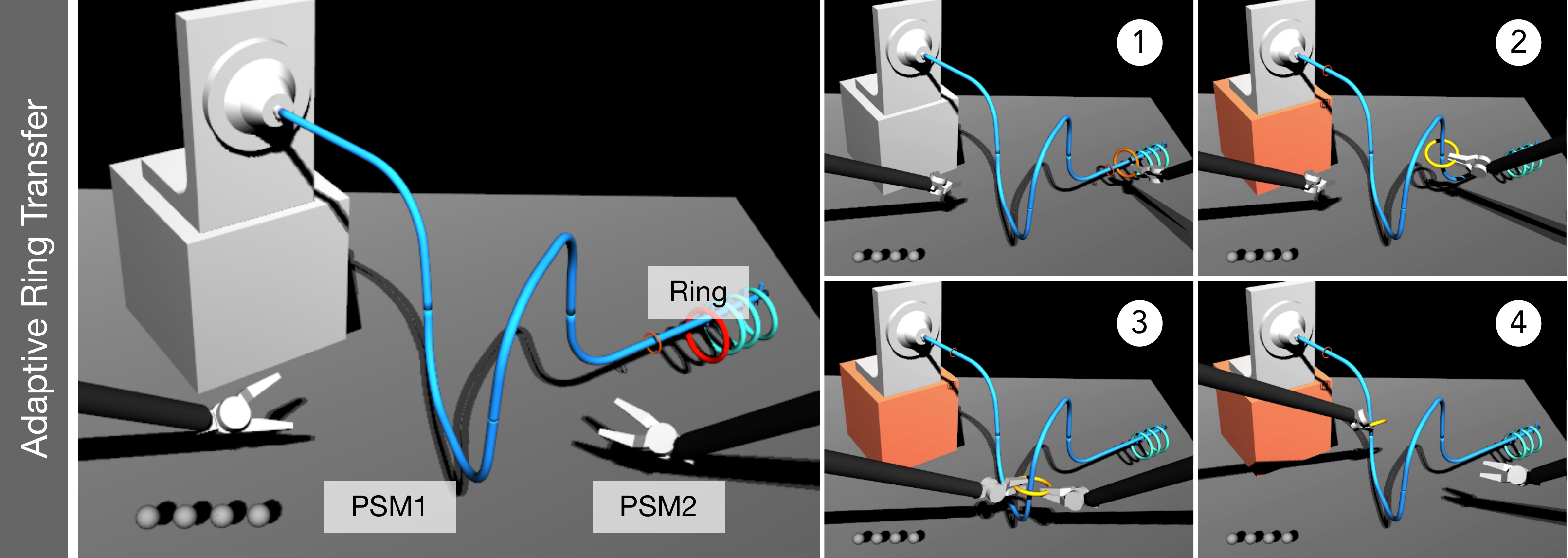}
      \caption{Snapshots of unlabeled VR ring transfer task (target domain dataset). Subfigures show the involved main gestures: (1) reach out the right arm to grasp the ring, (2) transfer the ring along the wire, (3) switch the ring from right to left, and (4) release the ring. 16 subjects participated in 10 skill simulation sessions (8 repeated exercises in each session). During training, adaptive haptic assistance (wrench to the master manipulators) were generated to assist trainees in controlling the virtual patient side manipulators (PSMs)~\cite{enayati2018robotic}.}
      \label{fig_ringtransfer}
\end{figure*}

 \begin{figure}[tb]
      \centering
      \includegraphics[width= 1\linewidth,clip ,trim=15pt 0pt 40pt 5pt]{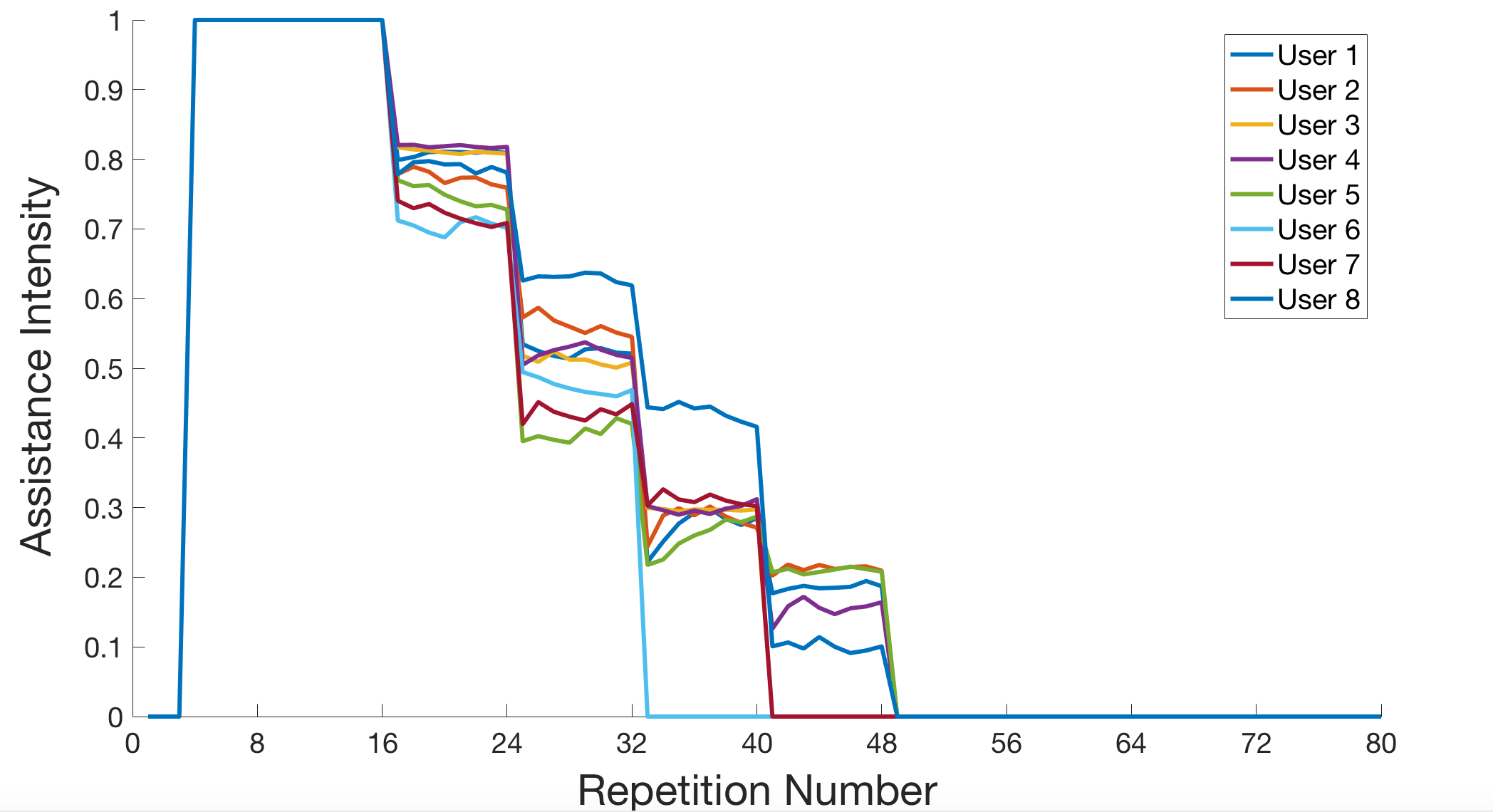}
      \caption{Intensity of haptic assistance for assisted subjects in every repetition of ring transfer task in the target domain.~\cite{enayati2018robotic}}
      \label{fig_assist_intensity}
\end{figure}

\begin{table*}[tb]
\renewcommand\arraystretch{1.4}
\renewcommand\tabcolsep{2.2pt}
\footnotesize
\caption{A comparison of datasets collected from independent surgical training tasks. Source: basic tasks of robot-assisted surgical training. Target: simulated ring transfer task with adaptive haptic assistance. dVSS denotes the da Vinci Surgical System and dVRK is the da Vinci Research Kit. MTM and PSM represent the master tool manipulator and patient side manipulator, respectively. }
\label{tab: dataset}
\centering
\begin{tabular}{ccccccccc}
\toprule[\heavyrulewidth]
\textbf{Category} &\textbf{Setup} & \textbf{Training Task} & \textbf{Controller} & \textbf{Assistance}  & \textbf{Skill Labels} & \textbf{Subject} & \textbf{Session} & \textbf{Sample Size}\\ 
\midrule
\multirow{3}{*}{Source} & \multirow{3}{*}{Dry Lab} & Suturing & \multirow{3}{*}{\parbox{2.5cm}{\centering dVSS MTMs\\dVSS PSMs}} & \multirow{3}{*}{N/A} & \multirow{3}{*}{\parbox{3cm}{\centering Labeled}}  & \multirow{3}{*}{8} & \multirow{3}{*}{-} & \multirow{3}{*}{120}\\
& & Needle Passing &&&&&& \\
& & Knot Tying &&&&&& \\
\midrule
\multirow{2}{*}{Target } & \multirow{2}{*}{Virtual Reality} & \multirow{2}{*}{\parbox{2cm}{\centering Ring Transfer}} & \multirow{2}{*}{\parbox{2.5cm}{\centering dVRK MTMs\\simulated PSMs}} & \multirow{2}{*}{\parbox{2cm}{\centering Adaptive\\Haptic Guidance}} &\multirow{2}{*}{Unlabeled} & \multirow{2}{*}{16} & \multirow{2}{*}{10} &  \multirow{2}{*}{1280} \\
&&&&&&&\\
\bottomrule[\heavyrulewidth]
\end{tabular}
\end{table*}

\section{Experiments}
\subsection{Datasets}
Two datasets in the field of robotic surgical training were used: the first dataset is denoted as the source domain, which contains labeled data with the skill annotations from basic surgical training tasks; the other is the target domain which contains unlabeled data collected from a VR simulation-based surgical training task that was characterized by the application of adaptive haptic guidance during its execution. Both datasets contain tool kinematics that were collected from \textit{daVinci} robotic surgical systems. The kinematics of two master tool manipulators (MTMs) and two corresponding patient-side manipulators (PSMs) were monitored and captured as multi-dimensional time-series. All kinematic variables were collected with 30 Hz sampling frequency. 
Specifically, the source domain data contains the joint kinematic variables from each robot manipulator, including Cartesian position (3), rotation matrix (9), linear velocity (3), angular velocity (3), and gripper angle (1), resulting 76-dimensional time-series per trial, while the target domain data contains all variables except for the measures of joint linear velocity, resulting 48-dimensional time-series per trial. To align with the data measurements in two datasets, we only considered the 48 common kinematic variables. The two datasets are described below and summarized in Table~\ref{tab: dataset}.

\subsubsection*{Source Domain Dataset}
The source domain dataset are labeled kinematics samples from a public-available technical skill dataset, namely the JHU-ISI Gesture and Skill Assessment Working Set (JIGSAWS)~\cite{gao2014jhu}. Subjects were required to perform three basic surgical training tasks: \textit{Suturing}, \textit{Needle Passing}, and \textit{Knot Tying}. Fig.~\ref{fig_jigsaws} shows the snapshots of the training tasks from JIGSAWS. Each surgical task was completed over 5 repetitions. Eight subjects in total with varying robotic surgical experience participated in the experiment, including four novice trainees (who practice on dVSS $<$ 10 hours), two intermediate trainees (with 10 to 100 hours of practice), and two expert surgeons (with $>$100 hours of practice). For simplicity, in this work we only consider two-class skill labels, namely, \textit{novice} and \textit{expert}, and the trainees with less than 100 hours are considered as novice, although our method can be adapted into multi-class settings with novice, intermediate and expert classes.

 \subsubsection*{Target Domain Dataset}
The target domain includes samples obtained from a scenario of personalized assistance-as-needed VR surgical training exercises~\cite{enayati2018robotic}. 
In the experiment, each subject performed a \textit{ring transfer} or \textit{steady hand task}: the trainee moves a ring along a curved wire pathway, while attempting to avoid the ring and wire making contact and keeping the ring’s plane perpendicular to the wire’s tangent. Fig.~\ref{fig_ringtransfer} shows some snapshots of the \textit{ring transfer} task. 
Sixteen subjects without medical background and with none to little experience in robotic teleoperation (12 males and 4 females, all right-handed) participated in the experiment and were randomly divided in two groups: an experimental (assisted) group and a null (non-assisted) group. The training task was carried out in 10 sessions over 5 days (2 sessions per day), where in each session subjects performed the same exercise over 8 repetitions.


For the group with robotic assistance (Assisted), trainees received an adaptive visco-elastic haptic guidance during the training session, that means forces and torques were applied to the MTMs in order to guide the trainee towards an ideal motion. The intensity of haptic guidance was adapted to a real-time measure of the subject’s performances according to the assistance-as-needed paradigm~\cite{radomski2008occupational,obayashi2014assist} in order to prevent cognitive overload at the beginning and slacking once trained. Consequently, the assistance intensity decreased all over the training as the trainees began learning how to perform the visuo-motor task. Fig.~\ref{fig_assist_intensity} shows the assistance intensity level provided for the group of subjects where each subject received full assistance after baseline assessment and the guidance was totally removed after 6 sessions. 

In contrast, trainees in the non-assisted group (Non-assisted) performed the task without receiving any robotic assistance throughout all the sessions. Details regarding the design of haptic assistance can be found in~\cite{enayati2018robotic}.

\subsection{Implementation Details}
Before feeding the motion kinematics into the network, we performed pre-processing for the raw data in two steps. First, we down-sampled data samples from both source and target domains by a factor of 30 (from the original sampling rate 30 Hz to 1 Hz). The purpose of down-sampling is to reduce unnecessary computational load. Then, normalization was also applied to each channel of all kinematics variables by subtracting the mean and dividing by the corresponding standard deviation. 

We implemented our methodology in Keras with Tensorflow 1.9 API as backend based on Python 3.6. All experiments in this study were remotely run on the UTSouthwestern BioHPC server and utilized a computing cluster equipped with a NVIDIA Tesla P100 GPU with 16 GB memory and 72 CPUs with 256 GB memory running a Linux kernel. 
To initialize the network, we first pre-trained the network on the source domain data given the known skill labels in JIGSAWS as ground-truth. We selected the best model as the one with the minimized loss on the validation set.
For training, we utilized the following hyper-parameters: 100 training epochs, stochastic gradient decent as chosen optimizer with initial learning rate as 0.001, first and second momentum of 0.9 and 0.999, and weight decay of $10^{-8}$. Before fitting the model in each epoch, the training set was randomly shuffled to achieve a robust process of model learning.

\subsection{Statistical Analysis of Pseudo Labeling in Target Domain}
Since no ground-truth skill labeling is currently available in the target domain dataset, we perform a statistical analysis in order to check for the effectiveness of pseudo labeling for extracting skill patterns. We hypothesize that the generated pseudo labels, given the skill knowledge has been generalized from the source to the target domain in the model, can provide meaningful significant results to reveal the motor skill learning pattern during a surgical training, and also be in align with prior studies that robotic assistance would help improve the proficiency of surgical training in general.  
We performed a two-way ANOVA analysis to determine the significance of skill probability between the training groups of subjects, the training sessions, and the interaction between the two. Similarly, another two-way ANOVA analysis was performed to check the significance differences between the training groups (Assisted and Non-assisted), the training stage (Pre-training and Post-training), and the interaction between the two.
A post-hoc Tukey comparison was used to compare different levels and highlight the significant effects within each group.

 \begin{figure*}[tb]
      \centering
      \includegraphics[width= 0.8\linewidth,clip ,trim=5pt 10pt 10pt 5pt]{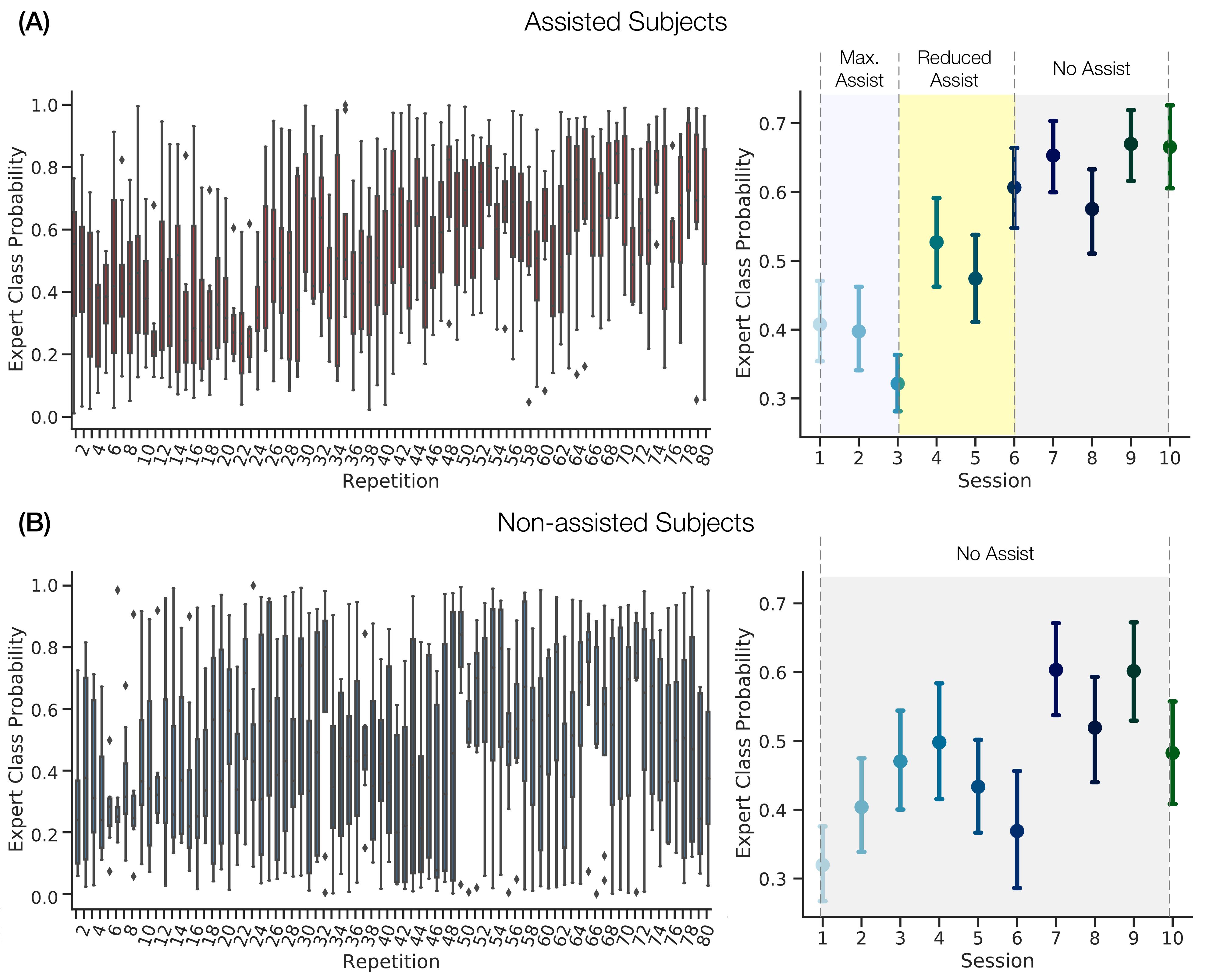}
      \caption{Average class probability of \textit{expert} in the target domain ring transferring with adaptive robotic assistance. (A) assisted subjects, (B) subjects without assistance. Left: the skill probability with respect to each repetition. Right: the probability in each session (8 repeated trials per session). The bar represents the standard deviation across the eight trials of each user in each session.}
      \label{rts_skill}
\end{figure*}

\section{Results \& Discussion}
To validate the effectiveness of our proposed approach for assessing skills, we first analyze the class probability of skills and model uncertainty when inferring skill labels for the unlabeled surgical training. For the simplicity of discussion, we only focus on the probability of the predicted \textit{expert} class. We first obtained the skill probability and uncertainty estimate of each trial and then averaged cross the eight trials in the corresponding session (8 repeated trials per session), respectively. 
In particular, we extract the motor learning curve from the output of skill predictions and investigate how the trainees' skills and model uncertainty varies as a result of robotic assistance and time when surgical training proceeds. Following that, we explore and compare the skill probability with task performance metrics that were measured in real-time during each trial of the surgical training. We hypothesize that the generated pseduo labels of the unlabeled data can inform the inherent learning curve in the surgical training and align with the results informed by task performance metrics. 

 \begin{figure}[tb]
      \centering
      \includegraphics[width= 0.9\linewidth,clip ,trim=5pt 5pt 10pt 5pt]{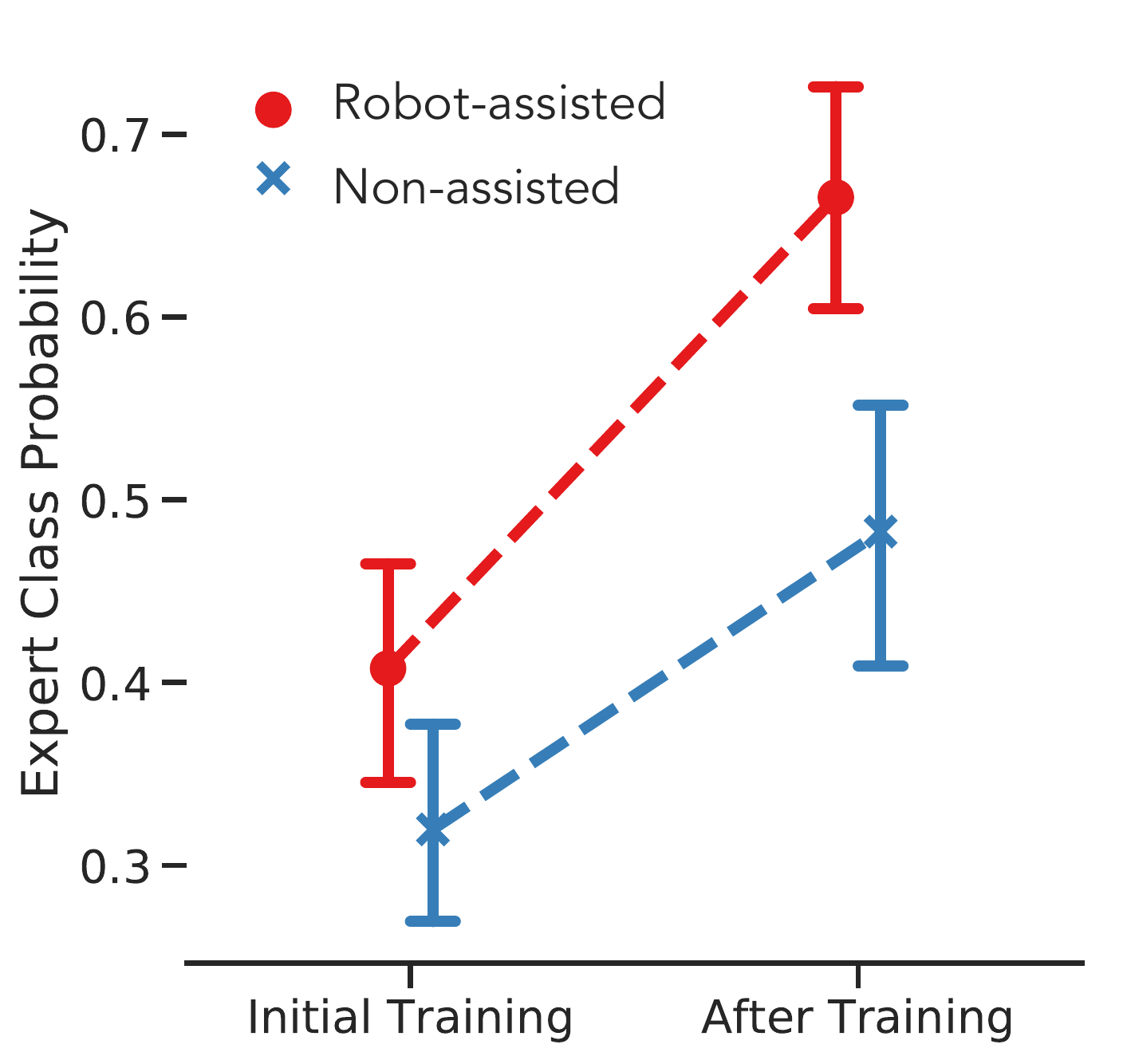}
      \caption{Class probability of \textit{expert} before (first session) and after (final session) training. Red denotes the subject group with robotic assistance and the blue represents the one without robotic assistance (non-assisted). The marker denotes the average value and the bar represents the standard deviation in each group. The slope of dotted line represents the motor learning rate. Overall, skills of both groups were significantly improved after training with higher \textit{expert} class probability ($p<0.05$), in align with performance metric measures, while the robot-assisted group shows a slightly larger learning rate.}
      \label{rts_skill_comparison}
\end{figure}

 \begin{figure}[tb]
      \centering
      \includegraphics[width= 1\linewidth,clip ,trim=5pt 10pt 10pt 5pt]{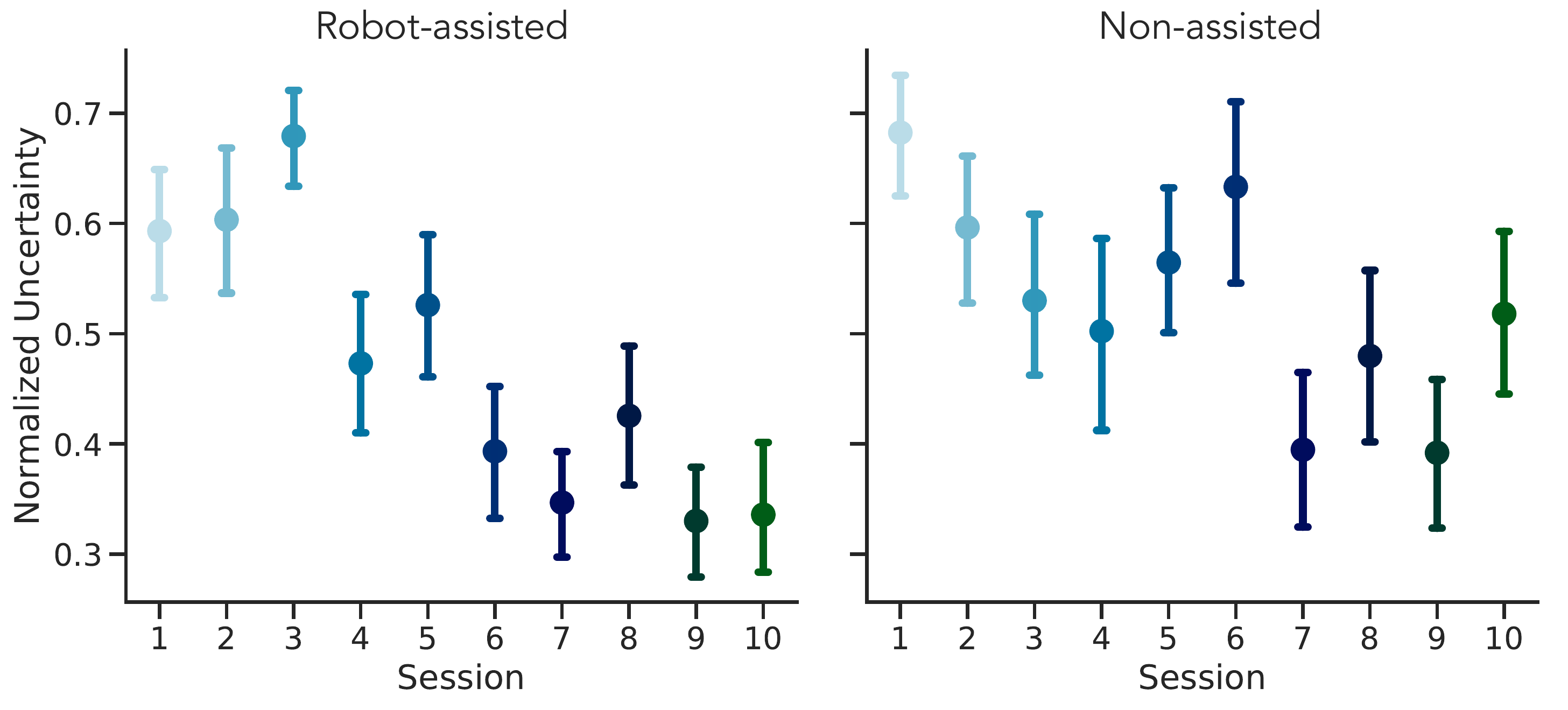}
      \caption{Uncertainty of predicted skill labels for the robot-assisted and non-assisted group in each session of surgical training. The bar represents the standard deviation across eight repetitions in a session.}
      \label{rts_uncertainty}
\end{figure}

\subsection{Skill Probability and Prediction Uncertainty}
Fig.~\ref{rts_skill} (A) and (B) present the \textit{expert} probability of robot-assisted and non-assisted subjects per repetition (left, averaged across corresponding group) and per session (right, averaged across all repeated trials), respectively. Table~\ref{tab_ANOVA} summarizes all the statistics.

\begin{table*}[tb]
\caption{Two-way ANOVA statistical analysis of the expert class probability within the subject group (Assisted/Non-assisted) and training group (Session, Pre/Post-Training). For simplicity, the post-hoc Tukey comparisons were considered within the group of subjects and the group of Pre/Post-training only. 
}
\label{tab_ANOVA}
\begin{center}  
\renewcommand\arraystretch{1.8}
\renewcommand\tabcolsep{11pt}
\begin{tabular}{cccccc}
\toprule[\heavyrulewidth]
 & \textbf{Subject Group} & \multicolumn{2}{c}{\textbf{Training Group}}  & \multicolumn{2}{c}{\textbf{Interaction}}\\
 \cmidrule(lr){2-2} \cmidrule(lr){3-4} \cmidrule(lr){5-6} 
& Assisted/Non-assisted & Session & Pre/Post-training & Assist$\times$Session & Assist$\times$Pre/Post-training\\
 \midrule
$p$ & $<0.001^*$ & $0.002^*$ & $<0.001^*$& $<0.001^*$ &  $0.692$ \\
\midrule
Tukey Comparison & Non-assisted$<$Assisted & - &  Pre-training$<$Post-training & - & - \\
\bottomrule[\heavyrulewidth]
\multicolumn{6}{r}{Asteroid $^*$ represents significance $p<0.05$}\\
\end{tabular}
\end{center}
\end{table*}

In general, skill learning of both assisted and non-assisted subjects was characterized by an increase of \textit{expert} class probability (or, equivalently, a decrease of \textit{beginner} class probability) as the surgical training proceeded. Significant difference was found between the assisted and non-assisted subjects on the skill probability, $p<0.05$.
Specifically, the subjects with robotic assistance were associated with considerably higher probability of expert after the initial three sessions. On the other hand, non-assisted subjects who did not receive any haptic assistance showed a continuous increase of technical skills within the first four sessions. However, non-assisted subjects had the larger variances of skill probability after the $4^{th}$ session. This result can be explained by the fact that the non-assisted group underwent a management for self-improvement as the surgical training repetitions proceeded. Nevertheless, it might be more difficult for trainees to pick up the most representative, \textit{expert}-relevant motion patterns without any assistance, and thus resulted in a less efficient learning outcome at the final stage of training. 
Importantly, the assisted subjects had a trend with continuous decrease of expert probability in the first three sessions (maximum haptic assistance). This result might be due to the fact that the advent of haptics may change the trainees' intended motion and consequently interfere their initial learning momentum, and thus trainees would need to adapt to the changes for an improvement. Nevertheless, the assisted subjects demonstrated a considerable improvement in their skills in and after the $4^{th}$ session. The differences of skill probability in the first three session and the $4^{th}$ session can be explained by the increased familiarity of subjects such that significant improvements are shown.

Fig.~\ref{rts_skill_comparison} compares the \textit{expert} class probability between the initial training (first session) and post training (last session). The figure highlights the motor learning outcome as the result of surgical training. As shown, either assisted or non-assisted subjects have significantly higher expert probability after the training compared to the initial training session. This result confirms that repeated training practices in general could help trainees in achieving a potentially higher expertise. Note that the steeper learning curve, as revealed by the dotted lines, indicates that the robotic assistance potentially increased the rate of learning for performing the complex motor task. Even though the assistance intensity was adaptively diminished after $3^{th}$ session and no assistance was provided after $6^{th}$ session, trainees were able to improve their technical skills more efficiently and to retain this improvement even after the assistance removal.

Fig.~\ref{rts_uncertainty} shows the averaged prediction uncertainties in each session of VR surgical training. The value of uncertainty is determined by the MC-dropout variational inference and reflects the level of predictive confidence when outputting a specific skill label given an input of motion kinematics. In general, surgical skills in the initial phase of surgical training are relatively more uncertain than the final stage. As the surgical training proceeds, the level of uncertainty for assessing skill decreases. The uncertainty could come from the variability between the observed motion data of trainees and the corresponding skill output. As the consequence of iterative practice, trainees could demonstrate distinguished motion patterns that provide a higher certainty of their skill levels, with or without robot assistance. We also note that non-assisted subjects are relatively more uncertain than the assisted subjects, especially at the end of surgical training. One interpretation is that between the assisted and non-assisted group, the motion profiles of assisted subjects could demonstrate more consistent signatures that are characteristic of high expertise; in contrast, subjects without robotic assistance might carry less distinguishable skill information in the motion that is hard to measure.

 \begin{figure*}[tb]
      \centering
      \includegraphics[width= 0.8\linewidth,clip ,trim=5pt 10pt 10pt 5pt]{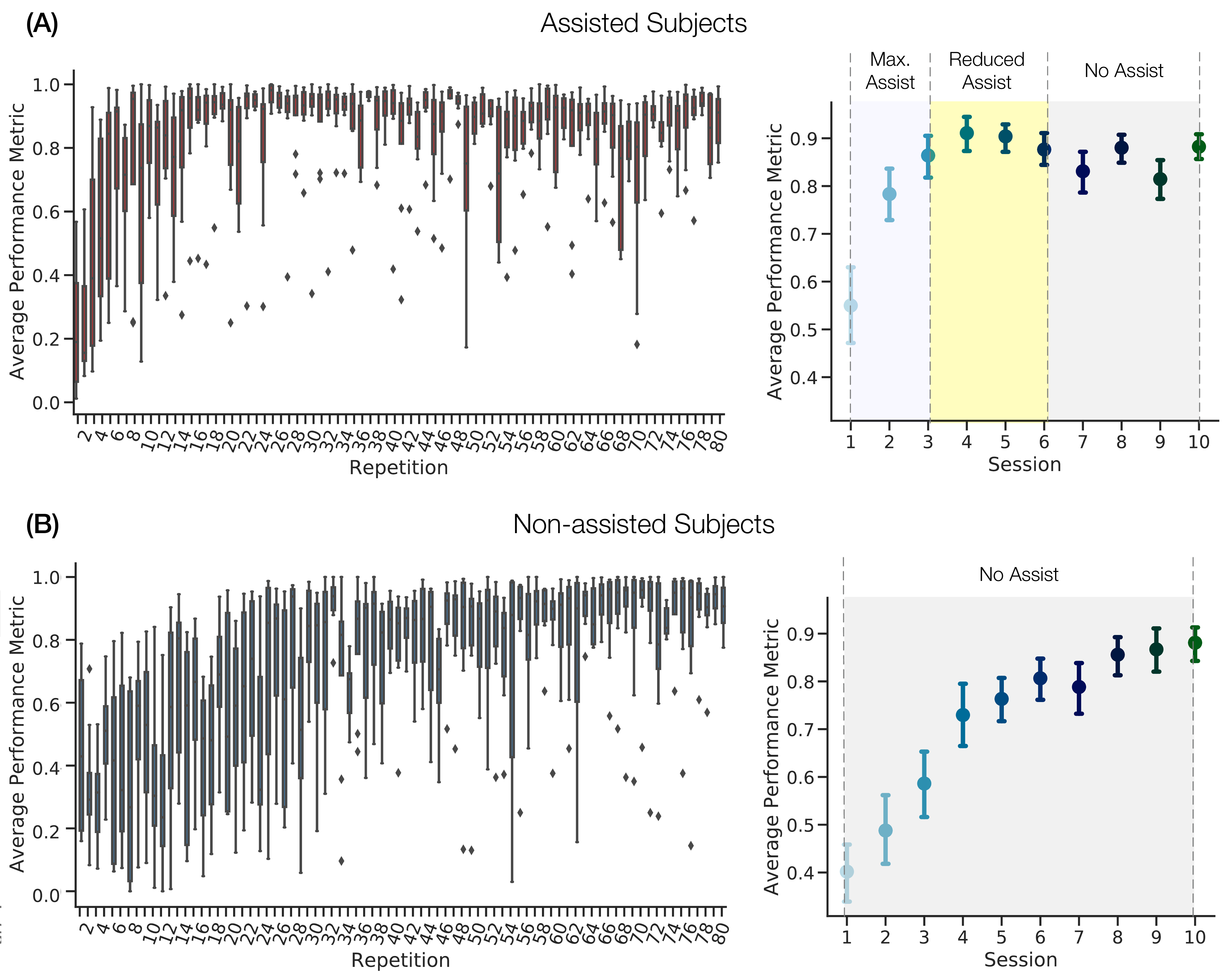}
      \caption{Overall task performance of (A) assisted subjects, (B) non-assisted subjects. The metric is calculated as the mean of three normalized measures of task performance: time, translational RMSE, and rotational RMSE, and indicates the overall performance of surgical operations. On the left, the performance of each repetition; on the right, the aggregated performance in each session. The error bar represents standard variance across eight repeated trials in each session.}
      \label{rts_perf}
\end{figure*}

\subsection{Compare to Task Performance Measures} 
In the absence of skill labels in the target domain dataset, we evaluated the validity of the generated pseudo labels by comparing their probabilities with the task performance metrics. These metrics, which include completion time, translational root mean square errors, and rotational root mean square errors, were used as a proxy measure of technical skills to indicate motor learning in the surgical training task. The task performance metrics were collected in each training trial and the average performance for all subjects is presented in Fig.~\ref{rts_perf}.
As shown, a clear learning curve of the trainees is captured by the task performance metrics. Consistent with our skill predictions, the non-assisted subjects improved their performance after multiple training sessions. Additionally, the assisted subjects showed better performance improvement due to the contribution of haptic assistance.

However, the evolution of the users' task performance metrics differs from the ones inferred from our skill predictions in terms of their slope and saturation. 
Specifically, the task performance metrics tend to be more sensitive to changes in assistance. We observed that, for the assisted subjects, the task performance measures rapidly improved when the maximum haptic assistance was activated, but showed relatively larger variances after the fourth session when the assistance intensity decreased. In contrast, as measured by the skill probability, the assisted subjects showed long-term persistence at the final stage of the training, indicating stable and consistent skill learning over time, even with a decrease in the intensity of haptic assistance. 
The discrepancy between the predicted skill and task performance metrics makes sense. The task performance metrics accurately reflect the direct impact of the physical assistance provided by the robotic system through its haptic guidance. The purpose of this guidance is to align the surgeon's current pose with the ideal one, resulting in optimal task performance metrics, by minimizing the translation and orientation errors. On the other hand, the predicted skills reflect a more complex motor learning process and changes in these skills may not be immediately reflected in the surgeon's performance. This could be considered an advantage of our approach for skill assessment, as it allows for the assessment of skills without any "assistance bias" in training scenarios that include robotic feedback or guidance.

\subsection{Limitations} 
This study focuses on adapting skill models from a fully-labeled dataset to a VR simulated training exercise and aims to investigate the transferability of skills across physical and virtual environment. The source data consists of only the labeled data of standardized training exercises, but its limited size may impact the generalizability of the skill models. 
To address this, future work could involve collecting a larger dataset for pretraining. 
The selected surgical training exercises from the source and target domains in our study are different but limited in term of the complexity. For future work, we plan to extend the analysis to include more complex tasks in boarder fields of surgical training, such as physical box trainers and wet-lab exercises~\cite{lahanas2016surgical}. We believe that a comparison of the skill models presented in our study and those measured on physical box trainers could help to provide insights into the training proficiency and outcomes across different environments~\cite{wang2019comparative,fathabadi2022box}.
Additionally, a more granular analysis of the skill models at surgical phase and task level might be helpful to reveal the corresponding surgeons' skills and workflow patterns~\cite{maier2018surgical,uemura2016procedural}. Furthermore, it would be interesting to explore the potential of this approach to generalize from technical to non-technical skill assessment in surgical training~\cite{nagyne2021non,agha2015role}. 


\section{Conclusion}
The present study aims to assess objective surgical skills across domains by adapting existing skill models from a labeled dataset to a virtual reality (VR) training exercise. Our method leverages the knowledge gained from basic robot-assisted surgical training exercises and enables a valid adaptation to unlabeled kinematic data.

To the best of our knowledge, this is the first study to generalize skill assessment to diverse surgical training exercises where full annotations are often difficult to obtain. We evaluated our method in a cross-domain surgical training task that employs adaptive assistance for trainees. Our approach is capable of learning domain-invariant features from both labeled and unlabeled data, and was able to construct learning curves throughout the training task.
The results showed that trainees using robotic assistance have significantly higher probabilities of becoming experts compared to those without assistance ($p<0.05$). This result is consistent with prior performance measures, indicating that proper robotic assistance can improve trainees' learning and speed of skill acquisition. The uncertainty in generating the corresponding skill label was also significantly lower with the assistance.

\section*{Acknowledgment}
This work is primarily supported by the National Science Foundation (NSF No. 1464432). This research was also supported in part by the computational resources provided by the BioHPC supercomputing facility (https://biohpc.swmed.edu), UT Southwestern Medical Center, TX, and the National Center for Advancing Translational Sciences of the National Institutes of Health under award UL1TR001105. The content is solely the responsibility of the authors and does not necessarily represent the official views of the NIH.

\bibliographystyle{IEEEtran}
\bibliography{main.bbl}

\end{document}